\documentclass[12pt, a4paper]{article}
\usepackage{lineno,hyperref}
\modulolinenumbers[5]

\usepackage[T1]{fontenc}
\usepackage[latin9]{inputenc}
\usepackage{color}
\usepackage{verbatim}
\usepackage{float}
\usepackage{amsmath}
\usepackage{graphicx}
\usepackage{url}
\makeatletter

\floatstyle{ruled}
\newfloat{algorithm}{tbp}{loa}
\providecommand{\algorithmname}{Algorithm}
\floatname{algorithm}{\protect\algorithmname}

\makeatother

\usepackage[english]{babel}

\begin{document}
%\begin{flushleft}
{\large
\textbf{A linear approach for sparse coding by a two-layer neural network \footnote{Published in Neurocomputing, Volume 149, Part C, Number 0, PP  1315 - 1323, Year 2015, DOI \url{http://dx.doi.org/10.1016/j.neucom.2014.08.066}, URL \url{http://www.sciencedirect.com/science/article/pii/S0925231214011254}}}
}
% Insert Author names, affiliations and corresponding author email.
\\
\vspace{5pt}
{\small
Alessandro Montalto$^{1,\ast}$, 
Giovanni Tessitore$^{2}$,
Roberto Prevete$^{3}$,
}
\\
\vspace{5pt}
{\scriptsize
\bf{1} Data Analysis Department, Ghent University, Ghent, Belgium
\\
\bf{2} Department of Physical Sciences, University of Naples Federico II
\\
\bf{3} DIETI, University of Naples Federico II
\\
$\ast$ E-mail: rprevete@unina.it
}
%\end{flushleft}
%\begin{frontmatter}

%\title{A linear approach for sparse coding by a two-layer neural network}

%\author[1]{Alessandro Montalto}
%\author[2]{Giovanni Tessitore}
%\author[3]{Roberto Prevete \corref{mycorrespondingauthor}}
%\address[1]{Data Analysis Department, Gent University}
%\address[2]{Department of Physical Sciences, University of Naples Federico II}
%\address[3]{DIETI, University of Naples Federico II}
%\ead{rprevete@unina.it}

\begin{abstract}
Many approaches to transform 
classification problems from non-linear to linear by feature transformation have been recently presented
in the literature. These notably include
sparse coding methods and deep neural networks.
However, many of these approaches require 
the repeated application of a learning process upon the presentation of unseen data input vectors, or else
involve the use of large numbers of parameters and hyper-parameters, which must 
be chosen through cross-validation, thus increasing running time dramatically.
In this paper, we propose and experimentally investigate a new approach for the purpose of overcoming  limitations of both kinds.  
The proposed approach makes use of a linear auto-associative network (called SCNN) with just one hidden
layer. The combination of this architecture with a specific error function to be minimized
enables one  to learn a linear encoder computing a sparse code which turns out to be as similar as possible to the sparse coding that one obtains by re-training the neural network.
Importantly, the linearity of SCNN and the choice of the error function allow one to achieve reduced running time
in the learning phase.
The proposed architecture is evaluated on the basis of two standard machine learning tasks. Its performances are compared with those of recently proposed non-linear auto-associative neural networks. 
The overall results suggest that linear encoders can be profitably
used to obtain sparse data representations in the context of machine learning
problems, provided that an appropriate error function is used during the learning phase. 
\end{abstract}

%\begin{keyword}
%neural networks \sep sparse coding \sep linear approach \sep encoder-decoder  
%\end{keyword}

%\end{frontmatter}

%\linenumbers

\section{Introduction}
\label{sec:introduction}
Various approaches to transform classification problems from non-linear to linear by a feature transformation have been recently investigated. These notably include sparse coding methods \cite{AharonEtAl_KSVD.2006, ranzatoEtAl_Sparse.2007,  mairalEtAl.2010, jenatton2010proximal, CurzioEtAl_PADDLE.2010} 
and deep neural networks \cite{hinton_fast.2006,bengio_learning.2009}. However, two remarkable limitations  usually affect these approaches: (i) most of the current sparse coding methods \cite[see, for example,]{AharonEtAl_KSVD.2006, mairalEtAl.2010, jenatton2010proximal, coatesEtAl_Analysis.2011} require the repeated application of some learning process in order to compute an input sparse representation, whenever the algorithm is fed with an input vector which was never used during the training phase; (ii) even though auto-associative neural networks enable one to overcome limitation (i), many of them, such as deep belief networks, present several levels of complexity or non-linearity. In fact, their learning methods usually involve a large number of parameters and hyper-parameters, such as the number of hidden units for more than one hidden layer, learning rates, momentum, weight decay, and so on. These parameters must be chosen through cross-validation, thus leading to a dramatic increase of running time. Interestingly, some authors \cite{coatesEtAl_Analysis.2011,dauphinAndBengio_Big.2013} suggest that simpler architectures, requiring a reduced number of parameters to be found,  may enable one to achieve state-of-the-art performance.
Following up this suggestion, the use of a relatively ``simple'' neural network approach, overcoming limitations (i) and (ii), is proposed here and experimentally investigated in terms of a trade-off between computational costs and performances.

The problem of overcoming limitations (i) and (ii) was addressed (a) by selecting an auto-associative neural network which enables one to learn a mapping between data and code space (the encoder) during the learning phase so that the learned mapping can be subsequently used on unseen data; and (b) by choosing just one linear hidden layer with identity as output function, jointly with an error function which allows one to learn at the same time both the encoder and the decoder by taking explicitly into account the contributions given by the two mappings. Given the linearity of the hidden layer and the selected error function, ``good'' learning rate parameters, ensuring linear rates of convergence in the minimization of error function, can be chosen in an explicit form without using cross-validation.

Points (a) and (b) guarantee that a reduced number of parameters must be determined during the learning phase: these are the number of hidden nodes and the sparsity parameter value which controls the extent to which the coding is sparse. On account of this fact, the present approach turns out to be computationally less expensive than deep neural networks and standard sparse coding approaches alike. Importantly, the present approach enables one to compute a linear encoder by means of which, during the test phase (on unseen data), one obtains a sparse code which is as similar as possible to the sparse code that one obtains by re-training the neural network. By applying a non-linear operation such as soft-thresholding or soft-max on this code one obtains a non-linear feature transformation.
The rest of the paper is organized as follows.The selected auto-associative neural network and its relation to previously proposed approaches are discussed in Section 2 and 3, respectively.
The experiment and its outcomes are presented in Section 4.
In particular,  in \ref{sub:Comparing-with-PCA} tests showing the capability of our approach to reproduce PCA behaviour are described and discussed, whereas the ability to obtain appropriate sparse data
representations enabling one to solve two standard machine learning tasks is evaluated in \ref{sub:Missing-Pixels}, \ref{sub:Digit-Classification} and \ref{sub:digit-classification-final}. Our approach is additionally compared with two auto-associative neural networks recently proposed in literature (\cite{zeng2010associative}, \cite{sivaram2010sparse} and with current results presented in the literature).
Finally, Section 5 is devoted to an analysis of experimental results and their significance.

\section{Background and related work}
\label{sec:related_works}

In the context of machine learning, the problem that we
are considering here is usually expressed in terms of a minimization
problem as follows:
\begin{align}
min_{\mathbf{U},\mathbf{D}}\Vert\mathbf{X}-\mathbf{U}\mathbf{D}^{T}\Vert_{F}^{2}+\lambda\Omega(\mathbf{U}) \nonumber \\
\text{Subject to }\Vert D^{j}\Vert_{2}^{2}\le1,\forall j
\label{eq:SC_Standard}
\end{align}

where $\mathbf{X}$ is a $N\times p$ matrix containing the $N$ $p$-dimensional
signals to be represented, $\mathbf{D}$ is a $p\times m$ matrix
containing the basis vectors (or atoms) column-wise arranged, $\mathbf{U}$
is a $N\times m$ matrix containing the sparse representations of
the signals in $\mathbf{X}$, $\Omega(\mathbf{U})$ is a
norm or quasi-norm regularizing the solutions of the minimization
problem, and the parameter $\lambda\ge0$ controls to what extent
the representations are regularized. The regularization term penalizes
the solutions containing many coefficients that are different from zero.

Importantly, approaches of this kind give rise to a crucial limitation: after the
learning phase, when an unseen data input vector $\mathbf{x}$ is considered, a minimization
process is again required to compute its sparse representation
$\mathbf{u}$. Sparse coding with auto-associative
neural networks overcomes this limitation, and several approaches based on this observation have
been accordingly proposed \cite[see, for example, ]{hintonAndSalakhutdinov_Reducing.2006,ranzatoEtAl_Unsupervised.2007,bengio_learning.2009,kavukcuogluEtAl_Fast.2010, wongEtAl_Robust.2013}.

In a nutshell, these approaches are based on an \emph{encoder-decoder}
architecture. The input $\mathbf{x}$ is fed into the encoder which
produces a feature vector $\mathbf{u}$, i.e., a code of $\mathbf{x}$.
In turn, the code $\mathbf{u}$ is fed as input into the decoder module
which reconstructs the input $\mathbf{x}$ from the code. Both encoder
and decoder are feed-forward neural networks which may present several
degrees of non-linearity. The encoder and the decoder are trained
so as to minimize the error between input $\mathbf{x}$ and reconstructed
input, with the proviso that the code $\mathbf{u}$ must satisfy certain 
given constraints in order to obtain a sparse code of $\mathbf{x}$. Sometimes
a further term is added in \eqref{eq:SC_Standard} so as to make the output of the encoder as similar as possible to the code $\mathbf{u}$ \cite[e.g.]{ranzatoEtAl_Sparse.2007,CurzioEtAl_PADDLE.2010}.
Such a term is added in our approach as well (see Section \ref{sec:SCNN}).
Importantly, as mentioned in the previous section, some authors
\cite{coatesEtAl_Analysis.2011,dauphinAndBengio_Big.2013} identified, 
as a drawback of many such architectures, their considerable complexity
and computational cost. Notably, the learning processes involved in these approaches usually 
require a large number of hyper-parameters such as learning rates, momentum,
weight decay, and so on, that must be chosen through cross-validation,
thus increasing running times dramatically. Furthermore, it is possible
to achieve state-of-the-art performance by means of simpler architectures
requiring the identification of a reduced number of parameters.

Recently proposed architectures
\cite{zeng2010associative,sivaram2010sparse}, that we now turn to examine,
involve non-linear auto-associative neural networks, called
respectively ASCNN and SAANN, and a single hidden layer. Both networks,
differently from our approach, can be used with a non-linear activation function
$\varphi$ (e.g., sigmoid for SAANN and tanh for ASCNN) on the hidden layer
units. 

In SAANN the error function expressed in \eqref{eq:sivaram-1} is used.
This function involves two terms. The first term is the standard
reconstruction error between the input signals $\mathbf{X}$ and the
reconstructed signals $\mathbf{U}\mathbf{D}^{T}$with $\mathbf{U}=\varphi(\mathbf{XC}^{T}$),
where $\mathbf{C}$ (a.k.a. projection dictionary) is the weight matrix
of the first weight layer of the network and $\mathbf{D}$ (a.k.a.
reconstruction dictionary) is the weight matrix of the second weight
layer of the network. The second term is a regularization term which
imposes sparse codes of the input. As one may readily note from \eqref{eq:sivaram-1}, 
the second term is a function
of the hidden units' output. Moreover the optimization algorithm simultaneously finds
the reconstruction dictionary $\mathbf{D}$ and the projection dictionary
$\mathbf{C}$ by a standard gradient descent technique.

\begin{equation}
E(\mbox{\textbf{D}},\mathbf{C})=\frac{1}{2}\|\mathbf{X}-\varphi(\mathbf{XC}^{T})\mathbf{D}^{T}\|_{F}^{2}+\lambda\sum_{n}\sum_{i}{log(1+u_{n i}{}^{2})}\label{eq:sivaram-1}
\end{equation}

In ASCNN the problems of finding the reconstruction
and the projection dictionaries are addressed separately by dividing
the autoassociative network into two subnetworks: \emph{top network}, and \emph{
bottom network}. The
optimization algorithm involves two basic steps:
\begin{enumerate}
\item Only the Bottom Network is considered. The input to the hidden units $\mathbf{Z}$
together with the reconstruction dictionary $\mathbf{D}$ are obtained
by minimizing the following error function:\\
\begin{equation}
E(\mathbf{Z},\mathbf{D})=\frac{1}{2}\|\mathbf{X}-\varphi(\mathbf{Z})\mathbf{D}^{T}\|_{F}^{2}+\lambda f\left(\frac{\mathbf{Z}}{\sigma}\right)\label{eq:zeng-1}
\end{equation}
\\
an alternating optimizations technique is used to find $\mathbf{Z}$
and $\mathbf{D}$. The first term in equation \eqref{eq:zeng-1} is the
usual reconstruction error, the second term is a regularization term
which imposes sparse solutions for the input codes. The sparseness function $f(\bullet)$
is chosen on the basis of the activation function $\varphi$ of the hidden units. 
A possible choice is $f(a) = log(1+a^{2})$ when the activation function corresponds either to $\tanh$ or to a linear function. $\lambda$ is a positive value
which controls the relevance of the regularization term.
\item Only the Top Network is considered. The projection dictionary $\mathbf{C}$
is retrieved by minimizing the following error function with respect
to $ $:\\
\begin{equation}
E(\mathbf{C})=\frac{1}{2}\|\mathbf{XC}^{T}-\mathbf{Z}\|_{F}^{2}
\end{equation}
leaving $\mathbf{Z}$  fixed.
\end{enumerate}
Finally the whole autoassociative network is considered in order to
achieve a fine tuning of network parameters ($\mathbf{Z}$, $\mathbf{C}$,
and $\mathbf{D}$) by minimizing the error function \eqref{eq:zeng-1}
and by taking into account that $\mathbf{Z}=\mathbf{XC}^{T}$.

Both networks, ASCNN and SAANN, are trained by means of the mini batch stochastic gradient
descent learning algorithm because the error function is differentiable
as the penalization term is differentiable. %

The possibility of identifying valid alternatives to non-linear approaches to sparse coding,
in the context of non-linear autoassociative networks with a single
hidden layer, is suggested by some classification which were successfully addressed on the basis of linear network approaches. Notably, 
linear approaches were successfully used to model the early stage responses
of the visual system. In the early stage 
visual information is a small number of simultaneously
active neurons among the much larger number of available neurons.
The first attempt to model this behaviour of the visual system is due to Olshausen
and Field (\cite{olshausen1996emergence}). The authors built a simple
single weight layer feed forward neural network (Sparsenet) where
the observed data $\mathbf{X}$ are a linear combination of top-bottom
basis vectors $\mathbf{D}$ and top-layer sparse responses $\mathbf{U}$.
The sparsity of the solution is obtained by minimizing an error function
composed of the standard sum-of-squares error and a regularization
term, which can be expressed as follows:

\begin{equation}
E(\mathbf{D},\mathbf{U})=\frac{1}{2}\parallel\mathbf{X}-\mathbf{U}\mathbf{D}^{T}\parallel_{F}^{2}+\lambda\sum_{i=1}^{m}g\left(\frac{\mathbf{U}_{i}}{\sigma}\right)\label{eq:sparseNet-1}
\end{equation}

where $g$ is a sparse inducing function, $\sigma$ is a scaling constant,
and $\lambda$ is the usual positive value which controls the relevance
of the regularization term.

\section{Sparse Coding Neural Network: a linear approach}
\label{sec:SCNN}
In order to investigate more systematically linear approaches, in the context of encoder-decoder architectures,
we introduce here:
\begin{itemize}
\item an auto-associative network with linear activation functions only;
\item an error function which allows one to learn at the same
time both projection and reconstruction dictionaries by taking
explicitly into account the contributions given by these two dictionaries;
in particular, we introduce a term in the error function which enables
one to obtain a sparse code which is as similar as possible to the output of
the encoder.
\item a \emph{hard sparse coding} approach, i.e., a limited number of values
different from zero, by means of a non-differentiable term in the
error function. We use \emph{hard sparse coding} also in view of the fact that it is
an efficient representation of biological network behaviours (\cite{rehn2007network}). 
\end{itemize}
Thus, we build a linear autoassociative neural network with two weight
layers. From now on, this network will be called \emph{Sparse Coding Neural Network}
(SCNN). During the learning phase SCNN can be regarded as being
formed by two independent sub-networks (see figure \eqref{Fig1_SCNN}):
a top network, T-SCNN, which includes both the projection dictionary
$\mathbf{C}$ and the SCNN hidden layer, and a bottom nework, B-SCNN,
which includes both the reconstruction dictionary $\mathbf{D}$ and
the SCNN output layer. In this phase, T-SCNN and B-SCNN are iteratively
and successively trained. This training process is fundamentally based
on two consecutive stages. First stage: both B-SCNN input signals,$\mathbf{U}$,
and the reconstruction dictionary, $\mathbf{D}$, are learned by considering
$\mathbf{X}$ as target values and by imposing a specific constraint
on $\mathbf{U}$ to obtain sparse input signals. Second stage:
the projection dictionary $\mathbf{C}$ is learned for T-SCNN by considering
$\mathbf{X}$ as input values and $\mathbf{U}$ as target values.
Consequently, SCNN is trained by minimizing the following global error
function:

\begin{figure}
\begin{centering}
\includegraphics[clip,scale=0.9]{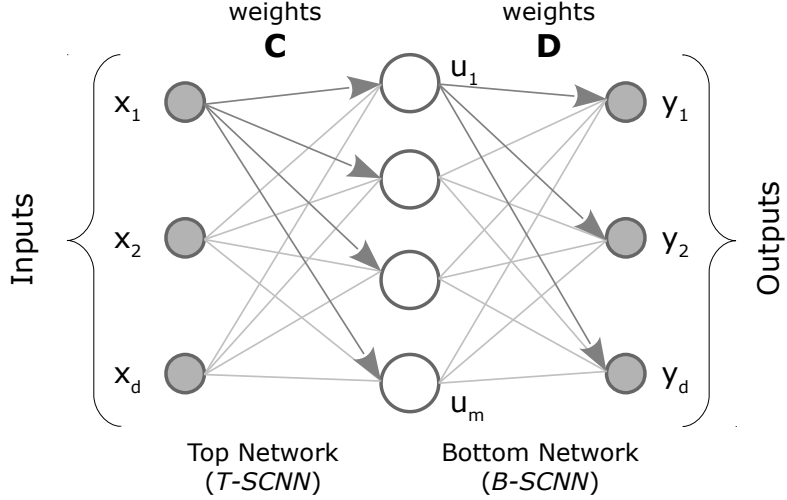}
\par\end{centering}

\caption{A graphical representation of the SCNN network. 
Top network (T-SCNN) includes
both the projection dictionary $\mathbf{C}$ and the hidden layer. Bottom network (B-SCNN)
includes both the reconstruction dictionary $\mathbf{D}$ and the SCNN output layer.}
\label{Fig1_SCNN}
\end{figure}

\begin{align}
E(\mathbf{D},\mathbf{C},\mathbf{U}) = & \frac{1}{p}
\sum\limits_{n=1}^{N}\sum\limits_{i=1}^{p}\left(x_{ni}- 
\sum\limits_{h=1}^{m}u_{nh}d_{ih}\right)^{2} + \nonumber \\
& \frac{1}{m} 
\sum\limits_{n=1}^{N}\sum\limits_{j=1}^{m}\left(u_{nj}- 
\sum\limits_{h=1}^{p}x_{nh}c_{jh}\right)^{2}+ \nonumber \\
&\frac{2\lambda}{m}\sum\limits_{n=1}^{N}\sum\limits_{h=1}^{m}|u_{nh}| \nonumber \\
\text{s.t.} \sum\limits_{h=1}^{p}\left(d_{hi}\right)^{2}\leq 1 & &
\label{eq:scnn-1}
\end{align}

where $x_{ni}$ is the $i$-th component of the $n$-th input signal, $u_{nh}$ is
the $h$-th component of the input code when the network is fed with the $n$-th 
input signal, $d_{ih}$ is the weight associated to the connection going from the $h$-th
hidden unit to the $i$-th output unit, and $c_{jh}$ is the weight associated to the connection 
going from the $h$-th input component to the $j$-th hidden unit. 
The first term forces the bottom network to reconstruct correctly
the input signals $\mathbf{X}$ on the basis of both $\mathbf{U}$
and $\mathbf{D}$, the second term in \eqref{eq:scnn-1} constrains
the solutions of the minimization problem to obtain B-SCNN input ($u_{nj}$)
as similar as possible to T-SCNN output ($\sum_{h=1}^{m}x_{nh}c_{jh}$),
and the last term imposes a sparse representation for $\mathbf{U}$.
Moreover, a quadratic constrain on the columns of $\mathbf{D}$ is
imposed to avoid $\mathbf{U}$ being the null matrix.

Let us now describe in more detail the proposed learning approach. Since the
error function (\eqref{eq:scnn-1}) is separately convex in each variable,
we proceed by iteratively minimizing with respect to B-SCNN
input variables $u_{ni}$, while keeping unchanged both $d_{ij}$ and
$c_{jk}$ (B-SCNN input update), then minimizing with respect to the reconstruction
weights $d_{ij}$ while keeping unchanged both $u_{ni}$ and $c_{jk}$
(B-SCNN weight update) and, finally, minimizing with respect to the projection
weights $c_{jk}$ while keeping unchanged both $d_{ij}$ and $u_{ni}$
(T-SCNN weight update). Importantly, the last term in \eqref{eq:scnn-1}
is not differentiable, and consequently it does not allow one to perform a
classical gradient descent learning algorithm. %

One can overcome this difficulty using proximal methods. Since
the partial derivatives with respect to the weights $d_{ij}$ and
$c_{jk}$ do not involve the non-differentiable term, these derivatives
can be computed following the standard back-propagation approach:

\begin{equation}
\frac{\partial E}{\partial d_{ij}}=-\frac{2}{d}\sum_{n=1}^{N}u_{nj}\left(x_{ni}-\sum_{h=1}^{m}d_{ih}u_{nh}\right)\label{eq:gradD-1}
\end{equation}

where $x_{ni}$ is taken as target for the bottom network , $d_{ih}$
is the weight from neuron $h$ to neuron $i$ of the bottom network,
and $u_{nh}$ is $h$-th component of the $n$-th B-SCNN input vector. 

\begin{equation}
\frac{\partial E}{\partial c_{jk}}=-\frac{2}{m}\sum_{n=1}^{N}x_{nk}\left(u_{nj}-\sum_{h=1}^{p}c_{jh}x_{nh}\right)\label{eq:gradC-1}
\end{equation}

where $u_{nj}$ is the $j$-th component of the $n$-th B-SCNN's input
vector taken as target for the top network, $c_{jh}$ is the weight
from neuron $h$ to neuron $j$ of the top network, and $x_{nh}$
is the $h$-th component of the $n$-th T-SCNN's input vector .

Thus the update of the weights $d_{ij}$ and $c_{jk}$ is obtained
on the basis of a standard gradient descend as follows:

\begin{equation}
d_{ij}^{s}=d_{ij}^{s-1}+\eta_{D}\frac{\partial E}{\partial d_{ij}}\label{eq:updateD-1}
\end{equation}

\begin{equation}
c_{jk}^{s}=c_{jk}^{s-1}+\eta_{C}\frac{\partial E}{\partial c_{jk}}\label{eq:updateC-1}
\end{equation}

where $s$ is the iteration step, and $\eta_{C}$, $\eta_{D}$ are
the learning rates. Moreover, to fulfill the quadratic constrain
in (\eqref{eq:scnn-1}) a projection operator on the unit ball is applied
on the columns of $\mathbf{D}$.

As mentioned above, since the last term in \eqref{eq:scnn-1} is not
differentiable, the update of $\mathbf{U}$ must be performed using a
proximal algorithm. In summary, a proximal algorithm minimizes a function
of type $E(\xi)=E_{1}(\xi)+E_{2}(\xi)$, where $E_{1}$ is convex
and differentiable, with Lipschitz continuous gradient, while $E_{2}$
is lower semicontinuous, convex and coercive. These assumptions on
$E_{1}$ and $E_{2}$ are necessary to ensure the existence of a solution.
The proximal algorithm is given by combining a projection operator
$(P)$ with a forward gradient descent step, as follows:

\begin{equation}
\xi^{s}=P\Bigl(\xi^{s-1}-\frac{1}{2\sigma}\nabla F(\xi^{s-1})\Bigr)\label{eq:proxStep-1}
\end{equation}

The step-size of the inner gradient descent is governed by the coefficient
$\sigma$, which can be fixed or adaptive.

In our case the \eqref{eq:scnn-1} can be minimized with respect to
$\mathbf{U}$ if considered as $E(\mathbf{U})=E_{1}(\mathbf{U})+E_{2}(\mathbf{U})$
where

\begin{equation}
E_{1}(\mathbf{U})=\frac{1}{p} \sum_{n=1}^{N}\sum_{i=1}^{p}\left(x_{ni}- \sum_{h=1}^{m}u_{nh}d_{ih}\right)^{2}+\frac{1}{m} \sum_{n=1}^{N}\sum_{j=1}^{m}\left(u_{nj}- \sum_{h=1}^{p}x_{nh}c_{jh}\right)^{2}
\nonumber
\end{equation}

and 

\begin{equation}
 E_{2}(\mathbf{U})=\frac{2\lambda}{m}\sum_{n=1}^{N}\sum_{h=1}^{m}|u_{nh}|
\nonumber
\end{equation}

The $E_{1}$'s gradient is

\begin{equation}
\frac{\partial E_{1}}{\partial u_{nk}}=-\frac{2}{p}\sum_{i=1}^{p}\left(x_{ni}- \sum_{h=1}^{m}u_{nh}d_{ih}\right)d_{ik}+\frac{2}{m} \left(u_{nk}- \sum_{h=1}^{p}x_{nh}c_{kh}\right)\label{eq:gradF-1}
\end{equation}

while the proximity operator corresponding to $P$ is the operator,
named \textit{soft thresholding}, defined as

\begin{equation}
P_{\lambda}(u_{nk})=sign(u_{nk})max\{|u_{nk}|-\lambda,0\}\label{eq:proxOperS-1}
\end{equation}

Now, replacing the \eqref{eq:gradF-1} and the \eqref{eq:proxOperS-1}
into the \eqref{eq:proxStep-1} and rearranging the terms, we obtain
the following updating rule for $\mathbf{U}$:
\begin{align}
u_{nk}^{s} &= P_{\lambda}\left(u_{nk}^{s-1}-\eta_{U}\frac{\partial E_{1}}{\partial u_{nk}}\right)  =  \nonumber \\
&P_{\lambda}\left(\left(1-\frac{2\eta_{U}}{m}\right)u_{nk}^{s-1} + \right. \nonumber \\ 
& \qquad  \left. 2\eta_{U}\left(\frac{1}{p}\sum_{i=1}^{p}\left(x_{ni}-
\sum_{h=1}^{m}u_{nh}^{s-1}d_{ih}\right)d_{ik} + \frac{1}{m} \sum_{h=1}^{p}x_{nh}c_{kh}\right)\right) 
\label{eq:updateU-1}
\end{align}

where $\eta_{U}$ is the learning rate.

Thus, the update expressed in \eqref{eq:updateU-1} plays a double role: it allows
one to find, on the one hand, sparse B-SCNN's input signals $u_{nk}$ able
to reconstruct the input data $\mathbf{X}$, and, on the other hand,
$u_{nk}$ values which can be \textquotedbl{}well approximated\textquotedbl{}
by the T-SCNN's output.

Furthermore, it is important to note that the choice of the parameters $\eta_{U}$,
$\eta_{C}$ and $\eta_{D}$  is crucial to achieve a minimum
of the error function \eqref{eq:scnn-1} as soon as possible. 
For an error function as $E = E_{1}+E_{2}$, the learning rate parameters
can be chosen on the basis of the Lipschitz constant of $\nabla E_{1}$
to speed up the iterative process. Hence, we 
set the parameters $\eta_{U}$, $\eta_{C}$ and $\eta_{D}$
as follows:

\begin{equation}
\label{eq:eta_U}
\eta_{U}=\frac{1}{2\|\frac{1}{d}\mathbf{D}^{T}\mathbf{D}+\frac{1}{m}\mathbf{I}\|}
\end{equation}

\begin{equation}
\label{eq:eta_D}
\eta_{D}=\frac{p}{2\|\mathbf{U}\mathbf{U}^{T}\|_{F}^{2}}
\end{equation}

\begin{equation}
\label{eq:eta_C}
\eta_{C}=\frac{m}{2\|\mathbf{X}\mathbf{X}^{T}\|_{F}^{2}}
\end{equation}

This choice of the learning rate values ensures linear rates of convergence in the minimization of the error function \cite{beck2009fast} and  convergence of both reconstruction and projection dictionary towards a minimizer \cite{beck2009fast}. In Algorithm \eqref{alg:scnn-alg} the pseudocode of the
learning process used to train SCNN is presented.

\begin{algorithm}
\textbf{INPUT}: input data $\mathbf{X}$.

\textbf{OUTPUT}: SCNN weight matrices $\mathbf{D}$ and $\mathbf{C}$.

\textbf{Inizialization}: initialize $\mathbf{U}^{0}$, $\mathbf{D}^{0}$
and $\mathbf{C}^{0}$ with random values taken from a uniform distribution
of values between $-1$ and $1$.

\textbf{Parameters}: setting the maximum number of external iterations
$T_{max}$, and the number of hidden neurons $m$.
\begin{enumerate}
\item $s\leftarrow1$
\item REPEAT 

\begin{enumerate}
\item \textbf{\textit{B-SCNN input update}}. B-SCNN inputs, $\mathbf{U}^{s}$,
at the current step $s$, are computed on the basis of $\mathbf{D}^{s-1}$
and $\mathbf{C}^{s-1}$ according to the equation \eqref{eq:updateU-1}
until convergence is reached;
\item \textbf{\textit{B-SCNN weigth update}}. B-SCNN weights, $\mathbf{D}^{s}$,
at the current step $s$, are computed on the basis of $\mathbf{U}^{s-1}$
and $\mathbf{C}^{s-1}$ according to \eqref{eq:gradD-1} until convergence
is reached;
\item \textbf{\textit{T-SCNN weigth update}}. T-SCNN weights, $\mathbf{C}^{s-1}$,
at the current step $s$, are are computed on the basis of $\mathbf{D}^{s-1}$
and $\mathbf{U}^{s-1}$ according to the \eqref{eq:gradC-1} until
convergence is reached;
\item $E(s)\leftarrow\frac{1}{p}\|\mathbf{X}-\mathbf{X}_{rec}^{s}\|_{F}^{2}$,
the error at the current step $s$ is computed as Frobenius norm between
the train set $\mathbf{X}$ and its reconstructed version $\mathbf{X}_{rec}^{s}=\mathbf{X}(\mathbf{C}^{s})^{T}(\mathbf{D}^{s})^{T}$
obtained as the result of the forward propagation of $\mathbf{X}$
through the global network SCNN;
\item $s\leftarrow s+1$;
\end{enumerate}
\item UNTIL $stop-condition(\mathbf{E},rtol,T_{max})$ is TRUE;
\end{enumerate}
\caption{\label{alg:scnn-alg}Learning algorithm for training SCNN network.}
\end{algorithm}

\section{Experiments}

Experiments of two different kinds were conducted. The first series of experiments
is aimed at evaluating the capability of the networks to reproduce
PCA behaviour (see Subsection \ref{sub:Comparing-with-PCA}).
In the second series of experiments,
we evaluated the performance of our approach
on the basis of two standard machine learning tasks. The first task is 
a \emph{missing-pixels} problem (see Subsection \ref{sub:Missing-Pixels}).
Here, we focused on evaluating the ability of 
the selected networks to obtain appropriate sparse data representations which enable one 
to solve the task. The second task concerns hand-written digit classification (see subsections \ref{sub:Digit-Classification}). In this latter task we first compared SCNN with the selected networks (see subsections \ref{sub:Digit-Classification}), and then 
we measured the error rate of our approach at varying the number of 
hidden nodes. Our results were compared with major extant approaches (see Subsection \ref{sub:digit-classification-final}). Moreover, we evaluated the effect of noise on the performance of SCNN.  
The main neural network parameters were set up as follows:
\begin{itemize}
	\item initialization of the weights. The initial weights were randomly chosen in the set $[-1,1]$;
	\item learning rates were chosen for SCNN to be the Lipschitz constants. The maximum number of iteration were always fixed to: $1000$ for step (a), $500$ for step (b), $500$ for step (c), and $50$ for external loop (see \ref{alg:scnn-alg});
	\item learning rate and maximum number of iterations for ASCNN and SAANN have been chosen in accordance with \cite{zeng2010associative} and \cite{sivaram2010sparse}, respectively;
	\item threshold was selected as stop condition. We heuristically tried different choices for all methods, and best results are reported;
	\item $\lambda$ was chosen according to what is specified in each experiment.
\end{itemize}

\subsection{Comparing SCNN with PCA}
\label{sub:Comparing-with-PCA}

%The first experiment concerns the capability of the networks to reproduce
%the PCA behaviour. 
Our approach is similar to a standard linear auto-associative network when one sets $\lambda=0$. However, in this case our error function is different from that used by linear auto-associative networks to reproduce PCA results. Hence, it is not obvious that the proposed method has solutions equal to PCA when $\lambda$ is $0$. Thus, the experiments in this Section are aimed at experimentally verifying whether our approach gives rise to a behaviour which is similar to PCA as worst case.
For this reason the networks were trained by
setting the sparsity parameter $\lambda = 0$. To evaluate the networks'
performance we built a training set extracting 2000 patches (small
parts of the whole image) of size $d = 8 \times 8$ pixels from the
Berkeley segmentation database of natural images \cite{martinEtAl_Database_2001},
which contains a high variability of scenes. We set the number of
hidden nodes equal to the principal components (10, 30, 50). Note
that the number of hidden nodes is always less than $d = 64$, i.e.,
the maximum number of the principal components available. In this
experimental setting the lowest reconstruction error is given by the
PCA.

\begin{figure}
\centering{\includegraphics[trim=2.6cm 1.5cm 2.8cm 1.1cm, clip=true,width=1\textwidth]{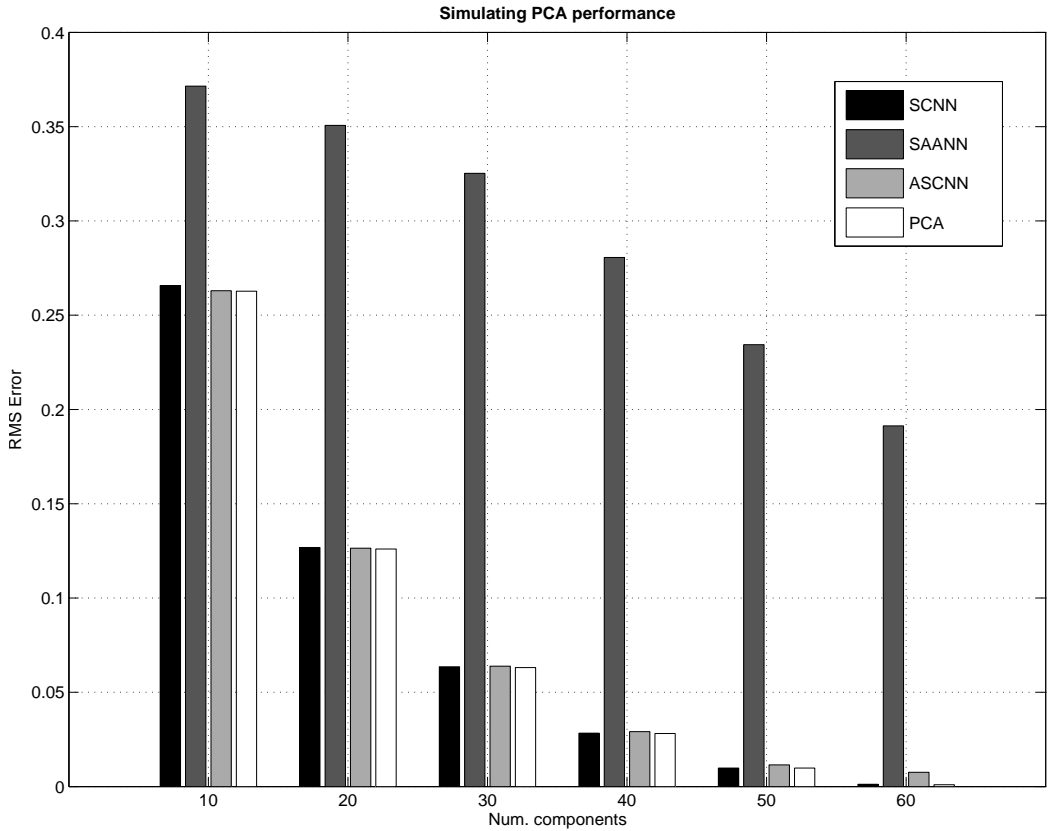}} 
\caption{\label{fig:simulatingPCA}The figure shows 
the values of the RMS Error
versus the number of principal
components. Note that the number of hidden nodes of the
considered neural networks is equal to the number of principal components. 
The standard deviation is not shown because,
for all methods, it is less than 0.02.}
\end{figure}

We trained the networks and, then, evaluated the reconstruction error
$20$ times. Each time we projected the input through the projection dictionary 
$\mathbf{C}$ and
reconstructed it through the reconstruction dictionary $\mathbf{D}$. 
We evaluated mean and standard
deviation of the reconstruction error; as shown in Figure \ref{fig:simulatingPCA} 
 SCNN and ASCNN can approximate PCA better than SAANN does. It is
worth noting that by increasing the principal components' number the
reconstruction error in SCNN and ASCNN decreases more than the SAANN's
reconstruction error. The reconstruction error was computed as the
Root Mean Square (RMS) error between original and reconstructed data.

\subsection{Missing Pixels}
\label{sub:Missing-Pixels}

\begin{figure}
\centering{}\includegraphics[trim=2.1cm 1.5cm 1.8cm 1.2cm, clip=true,width=1\textwidth]{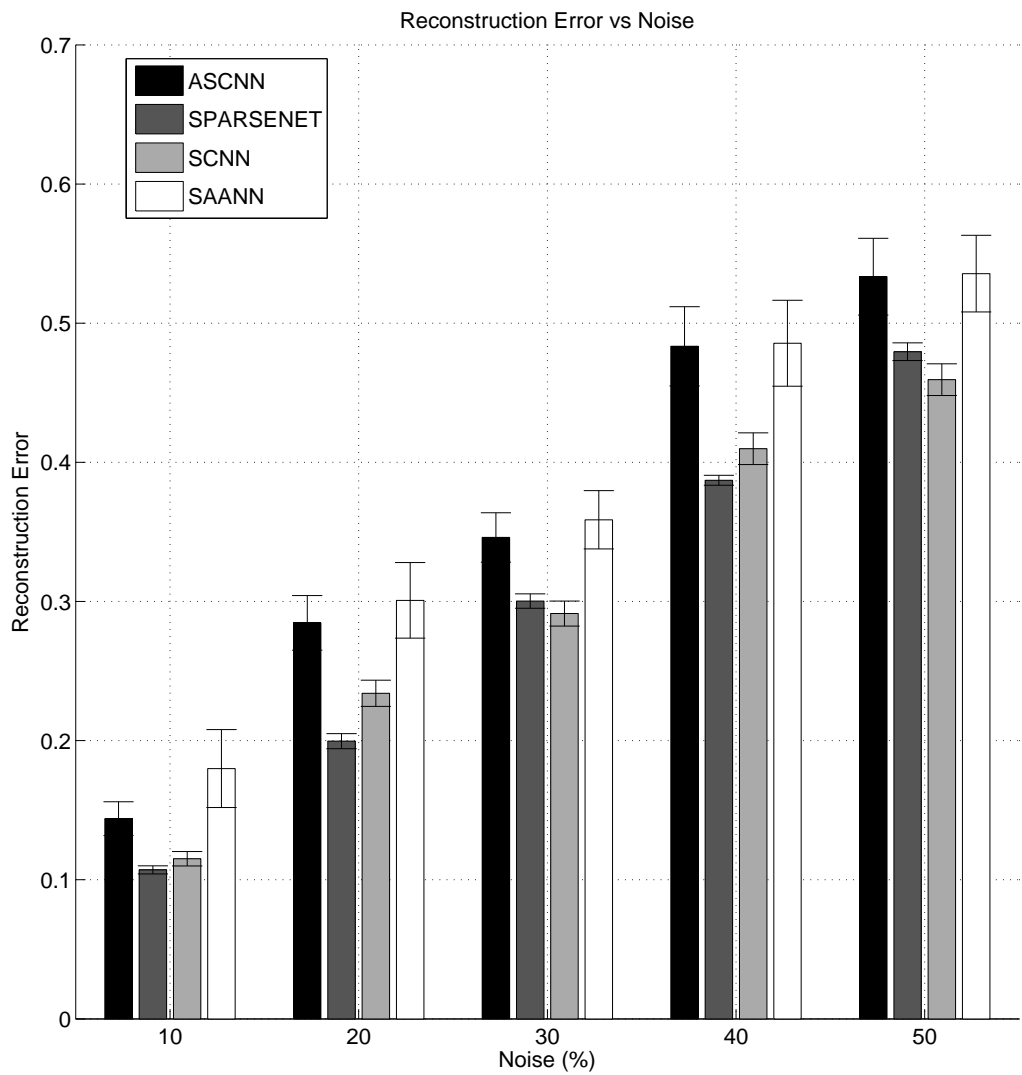}\caption{\label{fig:recErr} Reconstruction Error 
versus noise. Noise values correspond to $10\%$, $20\%$, $30\%$, $40\%$ and $50\%$ of missing pixels.}
\end{figure}

In this test we compared our approach with ASCNN, SAANN and Sparsenet on a well
known machine learning problem called \emph{missing-pixels}. 
We aim to reconstruct \emph{corrupted }(that is, setting randomly to zero
a certain amount of pixels) patches from a test set after having trained
the networks on a training set of \emph{non-corrupted} patches \cite{jenatton2010proximal}.
We extracted $4000$ patches of size $p=8\times8$ pixels from the Berkeley 
segmentation dataset of natural images \cite{martinEtAl_Database_2001}. We then split 
this dataset into a training set
%$\mathbf{X}_{tr}$ 
and a validation set 
%$\mathbf{X}_{val}$ 
of size $2000$ and $2000$ patches, respectively. All the patches are centered
so as to have zero mean. As test set, we used 2000 patches extracted
from the test images of the Berkeley dataset which were never seen by the networks
before. All the patches of both validation and test set were 
corrupted by setting to zero the same percentage of pixels. In particular,
we chose five noise levels corresponding to $10\%$, $20\%$, $30\%$, $40\%$ and 
$50\%$ of missing pixels. 
The four different approaches (ASCNN, SAANN,
Sparsenet and SCNN), were applied on the training set using $40$ equispaced
values of $\lambda$ in the range $[0.01,20]$ for Sparsenet, and in the range
$[0.01,1]$ for the remaining networks. For each network the
best solution was chosen on the validation set by considering the minimum
reconstruction error. Finally, the performances of the networks were
computed on the test set. Note that for each approach the reconstructed
images for the test set were obtained using the projection dictionary
chosen during the validation phase, except for Sparsenet where
a learning process is again required. As shown in Figure \ref{fig:recErr}
SCNN is able to reconstruct a test image better than ASCNN and SAANN 
for all the noise levels, whereas its performances are comparable with those of
Sparsenet.

Moreover, we computed the sparsity values of the four networks on
the test set. We selected the solutions corresponding to the parameter
$\lambda$ which gave the best performance on the validation set with
noise equal to $10\%$, $20\%$, $30\%$, $40\%$ and $50\%$. 

Note that the sparsity value 
was defined as $1-\frac{1}{nm}\sum_{j}\sum_{i}\theta(u_{ij}-Th_S)$,
where $u_{ij}$ are the coefficients of the $j$-th signal of the test
set, $\theta(x)$ is the Heaviside's function, and $Th_S$ is a threshold
value ranging in $[0,0.5]$. This approach to computing sparsity values
was motivated by the fact that many coefficients might be near to zero, but
not exactly zero. Consequently, this choice enables one to achieve a better evaluation 
of the networks ability to produce sparse data representations. 

\begin{figure}
\centering{}\includegraphics[trim=2cm 1.5cm 2cm 1.2cm, clip=true, width=1\textwidth]{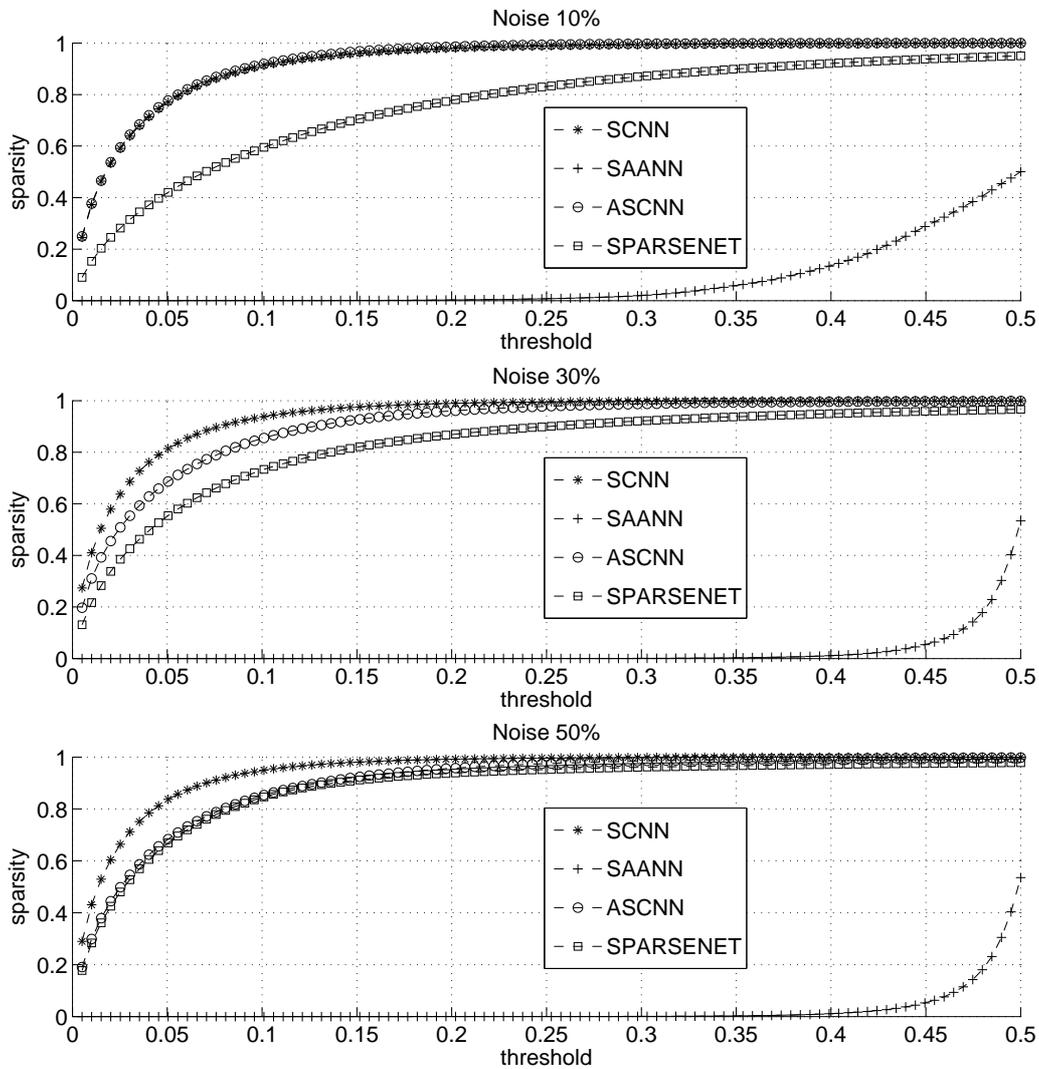}\caption{\label{fig:sparsity}Plot of the sparsity 
achieved on a test image versus the threshold $Th_S$ for different noise values.
The best performance is obtained by SCNN which turns out to be the more stable method with
respect to increasing percentages of noise. 
%SCNN achieves the same performances
%as ASCNN; its sparsity is even higher than Sparsenet that minimizes
%the error function during the test phase too.
}
\end{figure}

Figure \ref{fig:sparsity} shows these sparsity values against $Th_S$ values. 
Note that also in case of very low $Th_S$ values SCNN reaches high values of sparsity,
and for the different noise levels this ability is preserved. The performance of SCNN 
is equal to or better than the other selected methods.

\subsection{Hand-Written Digit Classification}

In this test we considered a hand-written digit classification problem.
The MNIST dataset \cite{lecunEtAl_Gradient.1998} is used here.
We extracted three datasets: a training
set $Digit_{tr}$, a validation set $Digit_{val}$ and a test set
$Digit_{T}$. Each image belonging to these datasets was transformed by re-mapping the
pixel range value into the interval $[-1,1]$. 
Each network was applied on the training
set using different values of $\lambda$.
For each $\lambda$ value a linear SVM was used as multi-class classifier
\cite{HastieEtAl_Elements.2001}. The classifier was trained with
the sparse image representations corresponding to the training set
$Digit_{Tr}$, and then it was fed with the sparse image representations
corresponding to the validation set $Digit_{val}$. For each network
the best solution was chosen on the validation set considering the
classification accuracy (defined as the ratio between correctly
classified images and the total number of images belonging to the
set). Finally, the performances of the networks were computed on the
test set $Digit_{T}$ considering the performance of the multi-class
classifier when it is fed with the sparse image representations obtained
by the previously selected projection dictionaries.
The test was organized into two phases. In the first phase, we compared our approach with ASCNN, SAANN and Sparsenet
on a reduced subset of the MNIST dataset (see Section \ref{sub:Digit-Classification}).
In a second phase, we measured the performance of our approach on the whole
MNIST dataset, in order to make a comparison possible with other methods presented in the literature (see Section \ref{sub:digit-classification-final}). 

\subsubsection{Comparing SCNN with ASCNN, SAANN and Sparsenet}
\label{sub:Digit-Classification}

Here SCNN is compared with ASCNN, SAANN and Sparsenet. 
The test is organized in three parts. 
In the first part, for a fixed size of the input, we computed classification accuracies 
on the basis of the data representations of the four networks at varying the number of hidden nodes, i.e., 
the number of atoms of the reconstruction dictionary.
Thus, we evaluated how the inner complexity of each neural network is reflected on its performances. Moreover,  we measured the computational times of each network during both learning and validation phase to compare the computational costs of our approach with the other methods.
In the second part of the experiment, for each approach we chose the neural architecture 
with a number of hidden nodes producing a ``satisfactory'' classification accuracy, and 
evaluated whether the ability to obtain data representations with high sparsity values was preserved in this case. 
In particular,
in order to get a better insight in the relation between accuracies and sparsity values, 
we computed first the area under the curve obtained by plotting the sparsity values versus the threshold $Th_S$, called \textit{sparsity area}, and then we computed the accuracies versus sparsity areas.
In the last part of the experiment, we evaluated to what extent each approach suffers from the ``curse 
of dimensionality'' \cite{bishop_Neural.1995}. Accordingly,
fixed the number of hidden nodes we computed the classification accuracies of the four networks  at 
varying sizes of the input.

\begin{figure}
\begin{centering}
\includegraphics[trim=1.8cm 1.5cm 2cm 1.2cm, clip=true, width=1\textwidth]{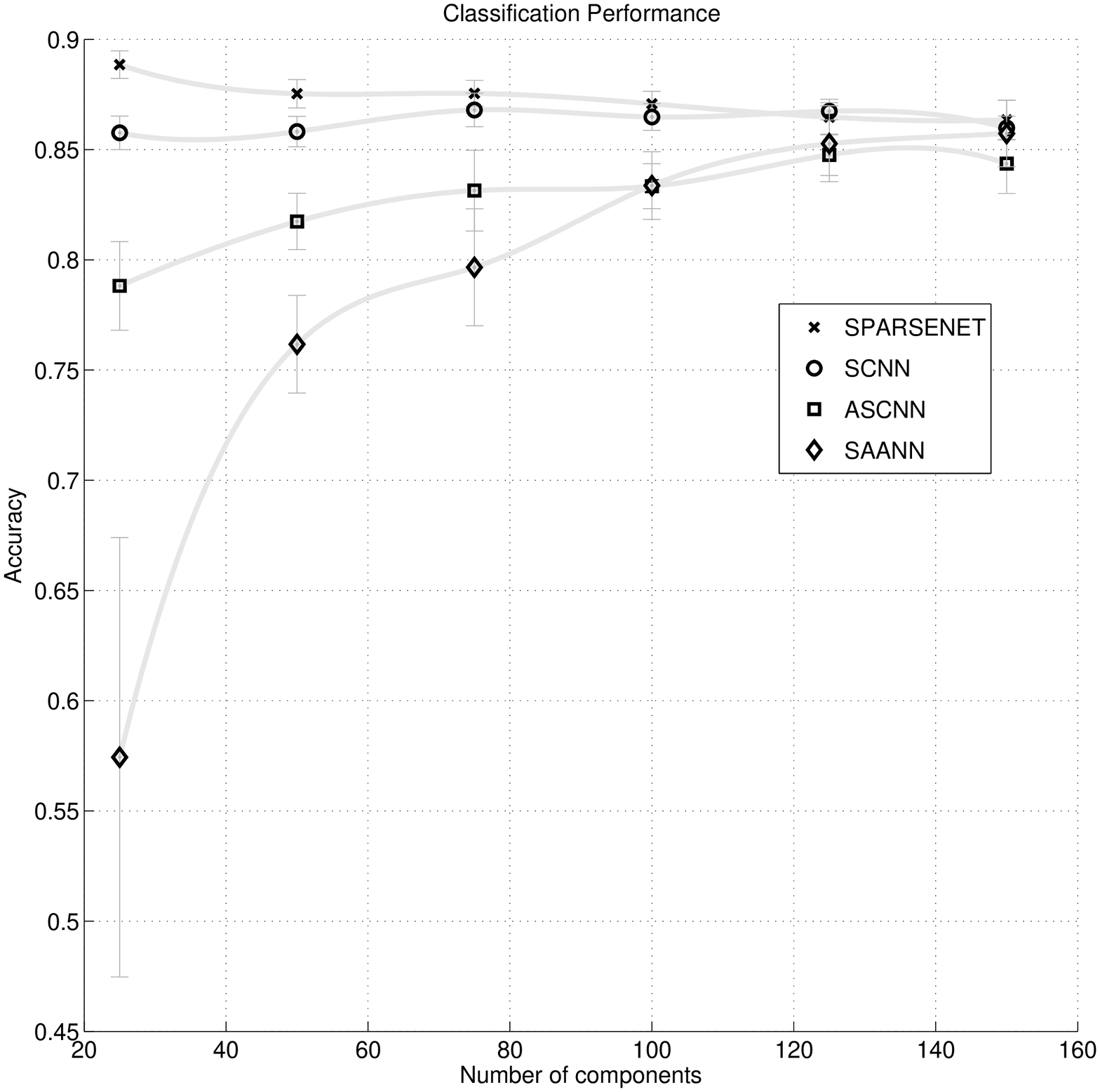}
\par\end{centering}

\caption{\label{fig:test1} Classification accuracy versus number of components.
SCNN achieves high accuracy even with $20$ hidden nodes while ASCNN and SAANN achieve the same performances as SCNN only when more than $100$ hidden nodes are allowed.}
\end{figure}

\begin{figure}
\begin{centering}
\includegraphics[trim=1.8cm 1.5cm 2cm 0.8cm, clip=true, width=1\textwidth]{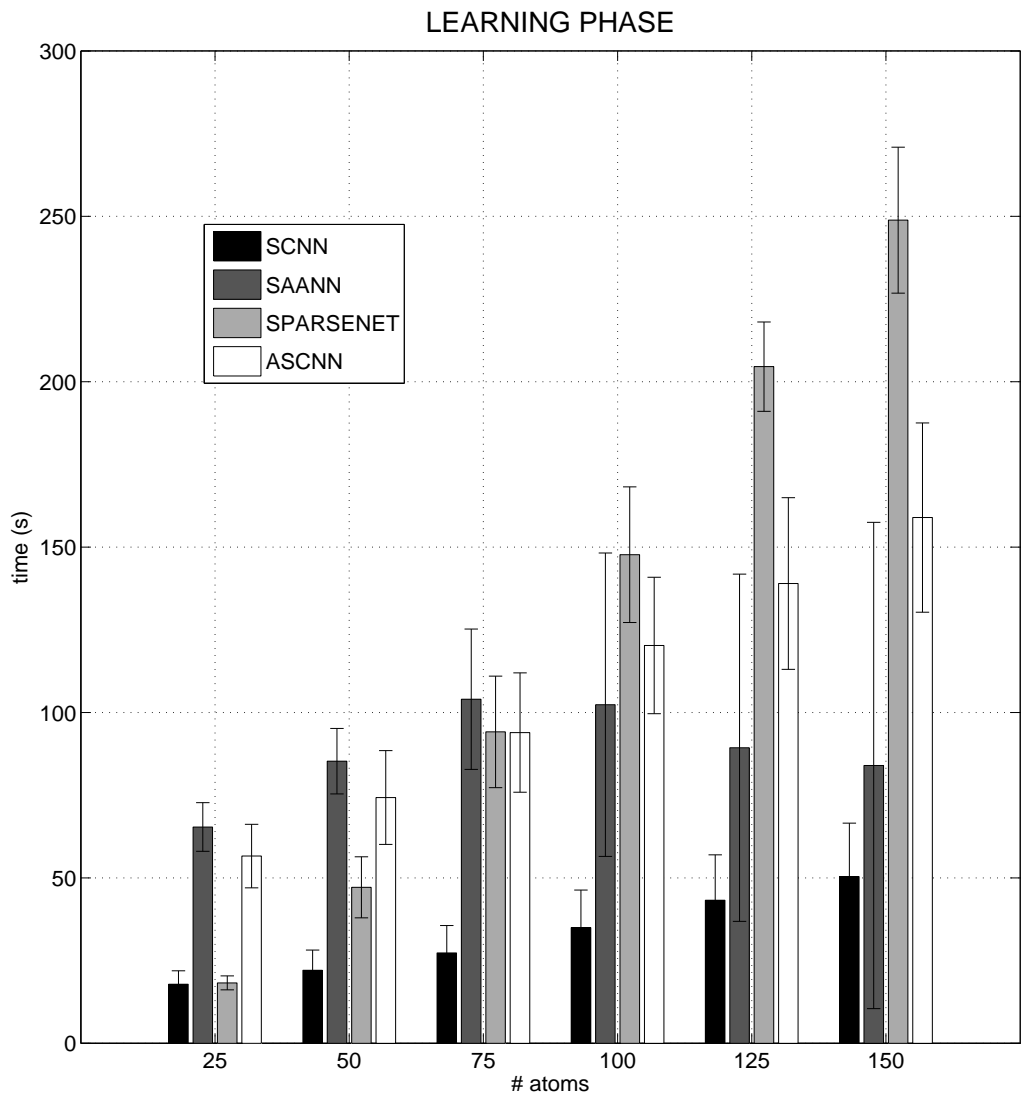}
\par\end{centering}

\caption{\label{fig:time1} Computational times during the learning phase for each neural network with respect to the dictionary dimension. The bars show the means of the computational times at varying the $\lambda$ parameter. The error bars represent the standard deviations.}
\end{figure}

\begin{figure}
\begin{centering}
\includegraphics[trim=1.6cm 0.8cm 2cm 0.8cm, clip=true, width=1\textwidth]{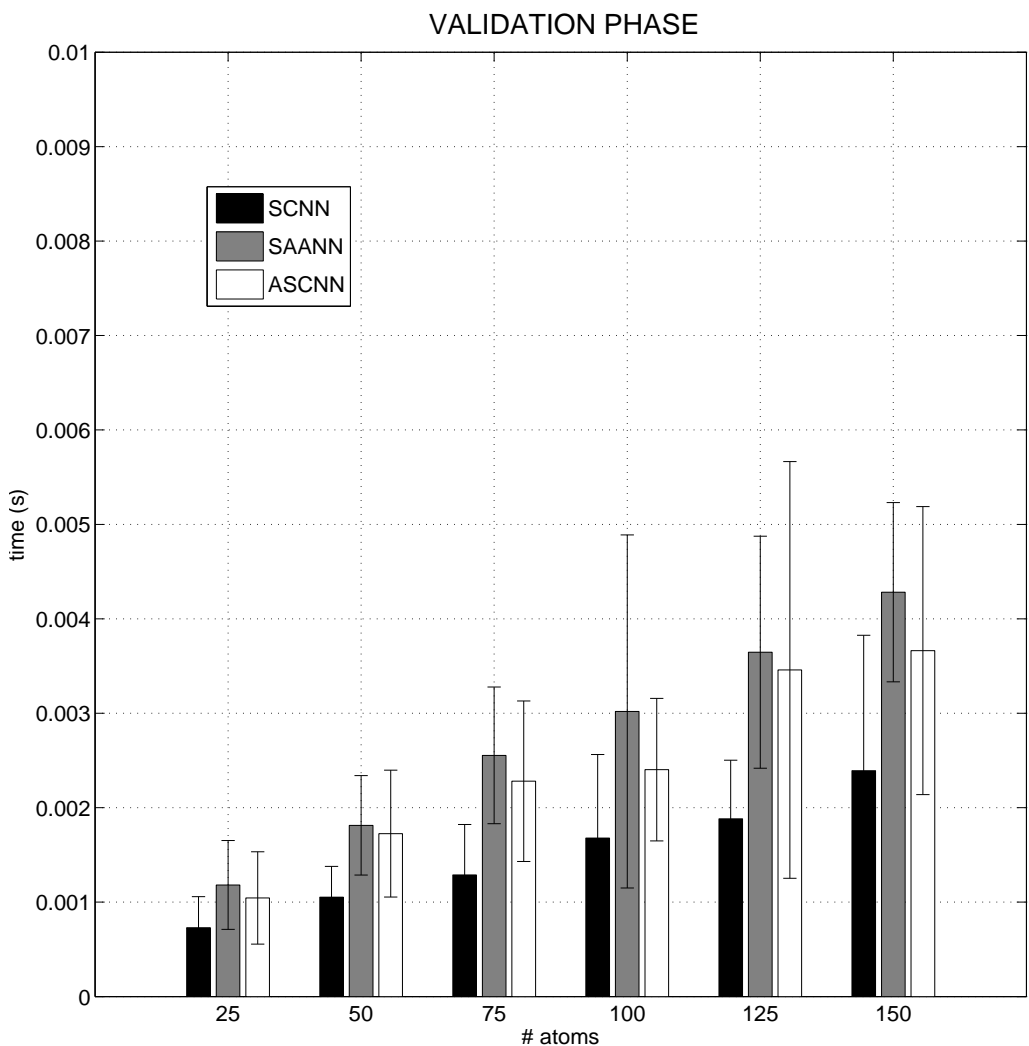}

\par\end{centering}

\caption{\label{fig:time2} Computational times during the validation phase for each neural network with respect to the dictionary dimension. Sparsenet computational times are not shown because they are an order of magnitude greater than those of the other neural networks. The bars show the means of the computational times at varying the $\lambda$ parameter. The error bars represent the standard deviations.}
\end{figure}

The sizes of the three datsets $Digit_{tr}$, $Digit_{val}$ and 
$Digit_{T}$ were chosen equal to $500$ digits. We chose $40$ equispaced values of $\lambda$ in the range $\ensuremath{[0.01,20]}$ for Sparsenet, and in the range $\ensuremath{[0.01,1]}$ for the other networks.
The
dictionary dimension in the first part of the test, Figure \ref{fig:test1}, was
varied from $25$ to $150$ with a step of $25$, and the digits 
were reshaped into a matrix of dimensions $14\times14$.
Notably, both Sparsenet and SCNN reach high values of classification
accuracies using a reduced number of atoms, and the classification accuracy
seems to be weakly dependent on this number, whereas
 ASCNN and SAANN need more than $100$ hidden nodes to reach 
performances that are comparable to those of Sparsenet and SCNN. In Figure \ref{fig:time1} and \ref{fig:time2} we show the means and the standard deviations of the computational times  at varying the $\lambda$ parameter for each dictionary dimension, and for all methods, during the learning phase and the validation phase, respectively. In Figure \ref{fig:time1} one can note that SCNN is uniformly faster than the other methods during the learning phase on all dictionary dimensions. In Figure \ref{fig:time2} Sparsenet computational times are not shown because they are an order of magnitude greater than those of the other methods. Also in this case, one can note that SCNN is faster than the other approaches for each dictionary dimension.

\begin{figure}
\begin{centering}
\includegraphics[trim=1.8cm 1.5cm 2.5cm 1.2cm, clip=true, width=1\textwidth]{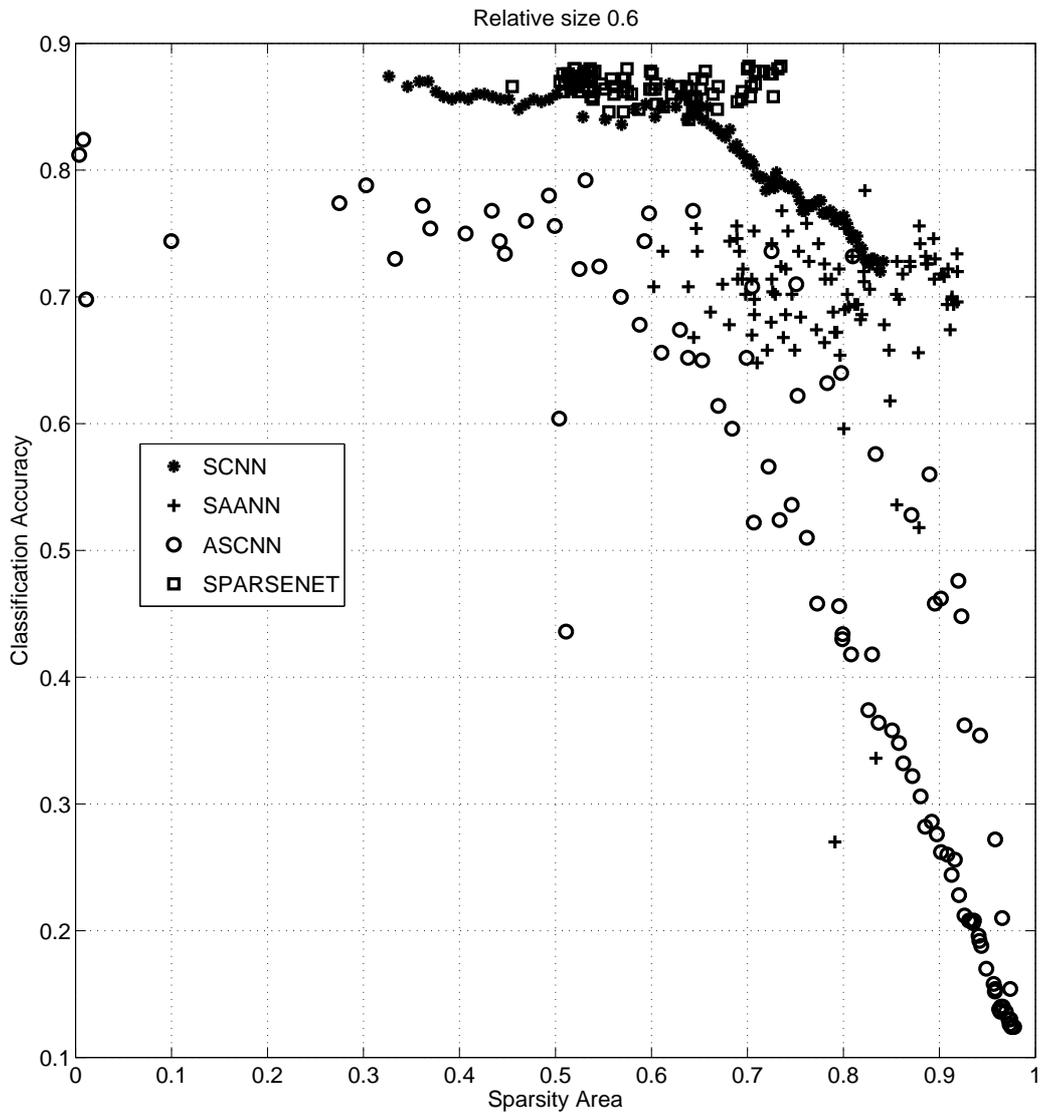}
\par\end{centering}

\caption{\label{fig:testAccSpars} Classification accuracy values versus sparsity areas.
SCNN seems to achieve the best compromise between sparsity area and accuracy in correspondence to sparsity area $\sim 0.8$ and accuracy $\sim 0.8$.}
\end{figure}

In the second part of the test, we fixed the size of the dictionary to $100$ and
computed classification accuracies and sparsity areas using $100$ 
equispaced values of the sparsity parameter $\lambda$.
In Figure \ref{fig:testAccSpars} classification accuracies versus
sparsity areas are showed for each network. 
Figure \ref{fig:testAccSpars} shows that the best accuracy values are obtained by
SCNN and Sparsenet. In particular, Sparsenet reaches the maximum
value (0.88) among the selected approaches. By means of SCNN one obtains high
accuracy values (more than 0.80) preserving high sparsity values in
terms of sparsity area. The other algorithms (ASCNN and
SAANN) exhibit high sparsity values but in connection with lower
values of classification accuracy only. More specifically, for high sparsity
values both SAANN and ASCNN reach accuracy values lower than 0.8.
On the whole, SCNN enables one to reach both high
accuracy values and high sparsity values.

\begin{figure}
\begin{centering}
\includegraphics[trim=2cm 1.5cm 3cm 1.2cm, clip=true, width=1\textwidth]{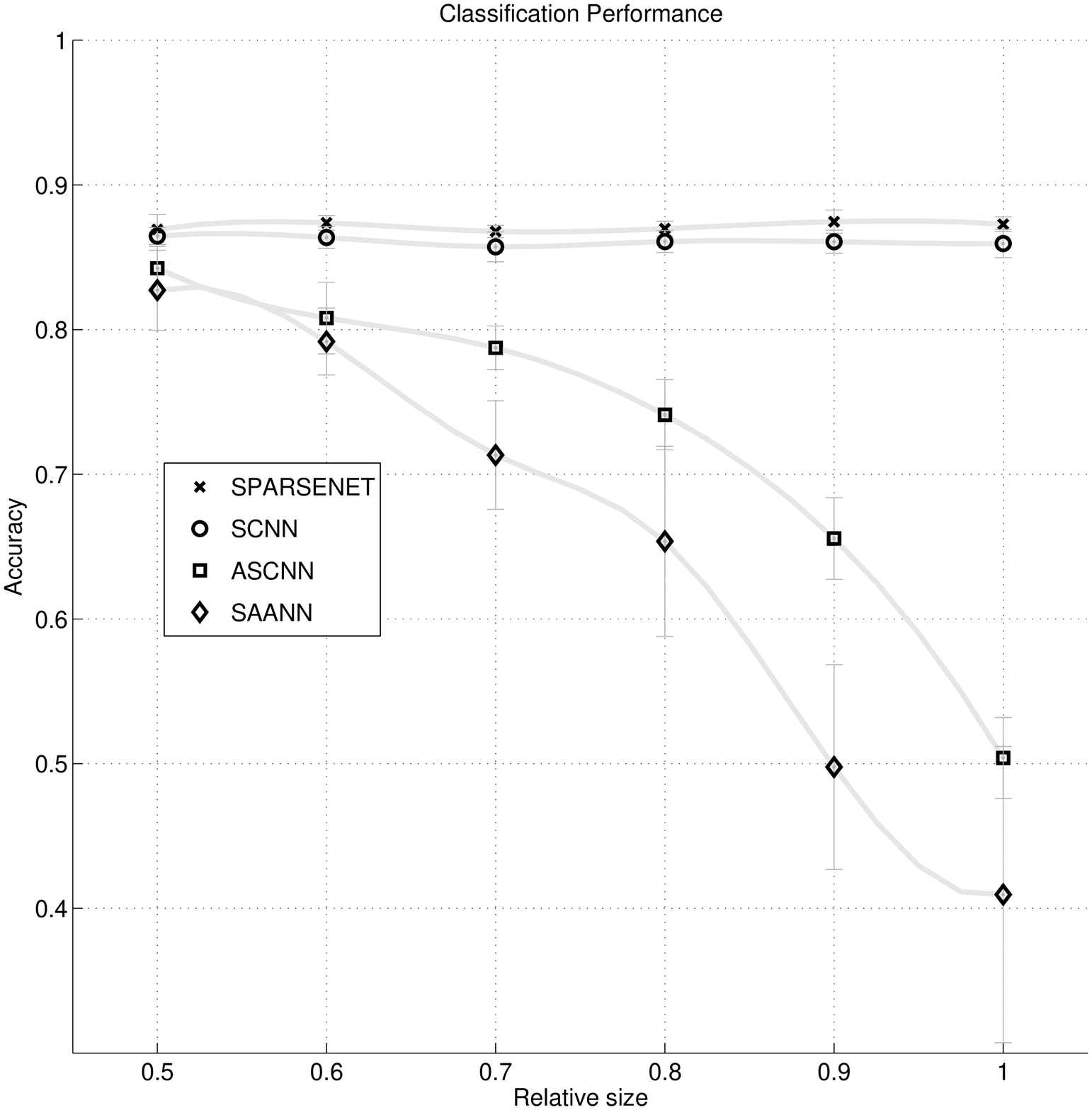}
\par\end{centering}

\caption{\label{fig:test2} Classification accuracy versus relative size. Leaving fixed the number of hidden nodes and the size of the training set while increasing the number of inputs by reshaping the digits, SCNN is very robust with respect to the ``course of dimensionality''.}

\end{figure}

In the last part of the test, Figure \ref{fig:test2},
we fixed both the size of the dictionary ($100$) and
the size of the training set, while digits were reshaped into
a matrix of dimensions from $14\times14$ to
$28\times28$. 
From Figure \ref{fig:test2}, one notices that the accuracy values reached 
by SCNN and Sparsenet do not seem to be affected by input size. By contrast,
ASCNN and SAANN turn out to be strongly dependent on input size.

%\begin{table}
%\begin{centering}
%\begin{tabular}{|c|c|c|c|c|}
%\hline 
%\# Components & SCNN & SAANN & ASCNN & PCA\tabularnewline
%\hline 
%\hline 
%50 & 70\% & 36\% & 38\% & 73\%\tabularnewline
%\hline 
%75 & 72\% & 44\% & 54\% & 76\%\tabularnewline
%\hline 
%\end{tabular}
%\par\end{centering}
%
%\caption{Table shows the percentage of correctly classified elements for each
%method. }
%\label{Tab_Classification}
%\end{table}

\subsubsection{SCNN Error Rate and Noise tolerance}
\label{sub:digit-classification-final}

In this phase the sizes of three datasets $Digit_{tr}$, $Digit_{val}$ and 
$Digit_{T}$ were of $50000$, $10000$ and $10000$ digits, respectively. 
The test was organized in two parts. 
In the first part, we measured the error rate of the linear multi-class classifier
on the sparse digit representations obtained by our approach at varying numbers of 
hidden nodes. The experimental setting was basically left unchanged with respect to the previously described setting. In particular,
we used $10$ equispaced values of $\lambda$ in the range $\ensuremath{[0.02,0.2]}$, and
a number of hidden node equal to $400, 800, 1200$ and $1600$. The sparse representations of  $Digit_{tr}$, 
$Digit_{val}$ and $Digit_{T}$ were obtained by applying the projection dictionary followed by
a soft-thresholding operation.

In the second part of the test, we evaluated the effect of noise on the performance of SCNN.
In particular we compared SCNN, raw-data and SCNN with a re-learning phase on
the test set (without using a projection dictionary) .  
We obtained $5$ versions of the  MNIST dataset by adding Gaussian white noise of mean $0$ and standard 
deviation
$\sigma = 0.02,0.04,0.06,0.08$. On each dataset we repeated the previously described experimental
setting with a number of hidden nodes equal to $400$.

The results show that our approach leads to performances that are quite good for this non-linear classification task,  insofar as they are
consistently better than those obtained by means of linear classifiers on raw images across various projection dictionary sizes. In 
addition, the best error rate was equal to $2.0\%$ which is better than or comparable to state-of-the-art 
results that are based on unsupervised feature learning plus linear classification without using additional image 
geometric information (see Table \ref{tab:digit-classification-all} and \ref{tab:digit-classification-scnn}). In 
particular, we note that the error rate of deep belief network  is very similar to that obtained by SCNN. Moreover, 
our approach seems to be little affected by noise (see Table \ref{tab:digit-classification-scnn-noise}).

\begin{table}
\begin{centering}
\begin{tabular}{|l|c|}
\hline 
 Methods & Error rate (\%) \tabularnewline
\hline \hline
Raw image $+$ Linear SVM & 12 \cite{yuEtAl_Nonlinear.2009} \tabularnewline
Sparse coding $+$ linear SVM & 2.02  \cite{cardosoEtAl_handwritten.2013} \tabularnewline
Deep Belief Network $+$ linear SVM & 1.90 \cite{yuEtAl_Nonlinear.2009}\tabularnewline
Stacked RBM network & 1.2 \cite{larochelleEtAl_Exploring.2009}\tabularnewline
Map transformation cascade $+$ linear SVM & 1.90 \cite{cardosoEtAl_handwritten.2013} \tabularnewline
Local Kernel smoothing & 3.48 \cite{yuEtAl_Nonlinear.2009} \tabularnewline
VQ coding $+$ linear SVM & 3.98 \cite{yuEtAl_Nonlinear.2009} \tabularnewline
Laplacian eigenmap $+$ linear SVM & 2.73 \cite{yuEtAl_Nonlinear.2009} \tabularnewline
Large Conv. Net, unsup. pretraining & 0.53 \cite{jarrettEtAl_best.2009}\tabularnewline
Local coordinate coding $+$ linear SVM & 1.90 \cite{yuEtAl_Nonlinear.2009} \tabularnewline
Human & 0.2 \cite{lecunEtAl_comparison.1995}\tabularnewline
\hline 
\end{tabular}
\caption{Error rates (\%) of MNIST classification with different methods. }
\label{tab:digit-classification-all}
\end{centering}
\end{table}

\begin{table}
\begin{centering}
\begin{tabular}{|l|c|c|c|c|}
\hline 
\# hidden units & 400 & 800 & 1200 & 1600 \tabularnewline
\hline \hline
SCNN $+$ linear SVM & 2.7 & 2.4 & 2.1 & 2.0 \tabularnewline 
\hline 
\end{tabular}
\caption{Error rates (\%) of MNIST classification by SCNN $+$ linear SVM against different number of hidden units.  }
\label{tab:digit-classification-scnn}
\end{centering}
\end{table}

\begin{table}
\begin{centering}
\begin{tabular}{|l|c|c|c|c|}
\hline 
Noise ($\sigma$) & 0.02 & 0.04 & 0.06 & 0.08 \tabularnewline
\hline \hline
Raw image $+$ linear SVM & 6.3 & 6.8 & 7.7 & 8.7 \tabularnewline
\hline 
SCNN $+$ linear SVM & 3.8 & 4.2 & 4.7 & 5.5 \tabularnewline
\hline 
SCNN with re-learning $+$ linear SVM & 4.9 & 4.8 & 5.7 & 6.9 \tabularnewline
\hline 
%SCNN $+$ LEARN $+$ ABS $+$ linear SVM & 4.3 & 4.6 & 5 & 6.7 \tabularnewline
%\hline 
\end{tabular}
\caption{Error rates (\%) of MNIST classification by SCNN $+$ linear SVM with a number of hidden units equal to $400$ against different noise values. These value are also compared with those obtained by raw images and
SCNN with a re-learning phase on the test set (without using a projection dictionary).}
\label{tab:digit-classification-scnn-noise}
\end{centering}
\end{table}

\section{Conclusions}

In this paper we showed that a linear two-layer neural netwok (SCNN)
can be profitably used to achieve sparse data representations for solving machine
learning problems. 

In our approach hidden layer linearity and the specific choice of error function allow one to explicitly define  learning rate values which ensure linear rates of convergence in the minimization of error function and  convergence of both reconstruction (decoder) and projection (encoder) dictionaries towards a minimizer \cite[see]{beck2009fast,combettes2006signal}.  Moreover, we have to set the number of units of just one hidden layer, unlike deep network approaches, and the linearity of the hidden layer allows one to perform a learning process which is computationally less expensive than a non-linear approach for a time constant. In this sense, we claim that our architecture is a comparatively \textquotedbl{} simple\textquotedbl{} one. 

SCNN reaches a very similar reconstruction error compared
to PCA as described in Section 4.1. Interestingly, by means of our approach, we obtained performances that are comparable
to or better than non-linear methods. 
More specifically, the experiments in Section \ref{sub:Missing-Pixels} and \ref{sub:Digit-Classification} show that our approach produces sparse data representations enabling one to solve standard machine learning problems. SCNN outperforms ASCNN and SAANN in all cases; and its performances are comparable to those of Sparsenet, which does not use an encoder to project unseen data to sparse code, but requires iteration of a learning process. In the hand-written digit classification problem (see Section \ref{sub:digit-classification-final}), our results are competitive with respect to state-of-the-art results that are based on unsupervised feature learning plus linear classification without using additional image geometric information. In addition, our approach seems to be little affected by noise.

It is worth noting, in connection with Section \ref{sub:Missing-Pixels}, that during the test
phase our approach (SCNN) reaches high sparsity values comparable
with or better than the values obtained by ASCNN and Sparsenet that make use of
non linear activation functions and a new learning process, respectively.
SAANN has sparsity values that are much lower than those obtained by all 
other methods. Figure 4 shows that the SCNN linear
encoder projects input data with a degree of sparsity greater
than or equal to that obtained by ASCNN and Sparsenet for different
noise levels. In connection with Section \ref{sub:Digit-Classification}, it is worth noting that
both SCNN and Sparsenet achieve high accuracy values (more than 0.80) preserving high sparsity 
values in terms of sparsity area. ASCNN and SAANN exhibit also high sparsity values 
but in correspondence of lower values of classification accuracy.
In the hand-written digit classification problem (see Section \ref{sub:digit-classification-final}, linear projection dictionary followed by soft-thresholding operator enable one to obtain an error rate equal to about $2\%$, that is,  a better value than that obtained by other approaches such as Local Kernel smoothing, VQ coding and Laplacian eigenmap. Furthermore, this value is comparable to standard sparse coding approaches and deep network (see Table \ref{tab:digit-classification-all} and \ref{tab:digit-classification-scnn}).

It is worth noting that the learning
process based on an alternate updating of the dictionaries $\mathbf{C}$
and $\mathbf{D}$ (see ASCNN and SCNN learning processes, sections
2 and 3) appears to lead to better results than the update involving
two dictionaries simultaneously (see SAANN learning process, Section
2). In fact, SAANN achieves in many cases worse results than SCNN and ASCNN
(see for example figures 2, 3, and 6).

Altogether, these results suggest that linear encoders can 
be used to obtain sparse data representations that are useful in the context of machine learning
problems, providing that an appropriate error function is used during the learning phase. 
In particular, on the basis of some suggestions made in
\cite{basso2011paddle,ranzatoEtAl_Sparse.2007} we introduced a term in the error function 
which enables one to obtain a
sparse code which is as similar as possible to the output of the encoder. 

\section*{Acknowledgements}
The authors wish to thank Guglielmo Tamburrini for his support
in preparing this manuscript.

\bibliographystyle{unsrt}
\bibliography{references}

\end{document}